\documentclass[11pt,a4paper]{article}
\usepackage[hyperref]{acl2021}
\usepackage{times}
\usepackage{latexsym}

\usepackage{times}  
\usepackage{helvet} 
\usepackage{courier}  
\usepackage{url}  
\usepackage{graphicx} 
\usepackage{bm}
\usepackage{xcolor}
\usepackage{microtype}
\usepackage{subfigure}
\usepackage{multirow}
\usepackage{multicol}
\usepackage{array}
\usepackage{amsfonts}
\usepackage{pifont}
\usepackage{algpseudocode}
\usepackage{booktabs}
\usepackage{amsmath}
\usepackage{hyperref}

\usepackage{amssymb}
\usepackage{tabularx}
\graphicspath{{fig/}}
\usepackage{float} 
\usepackage{bm}

\usepackage{enumitem}
\usepackage{soul}
\usepackage{arydshln}

\aclfinalcopy 
\def\aclpaperid{1901} 

\newcommand{\cmark}{\checkmark}
\newcommand{\xmark}{$\times$}

\newcommand{\todo}[1]{}
\renewcommand{\todo}[1]{{\color{red} TODO: {#1}}}

\title{\textsc{Mask-Align}: Self-Supervised Neural Word Alignment}

\author{
 Chi Chen$^{1,3,4}$,
 Maosong Sun$^{1,3,4,5}$, Yang Liu\thanks{\:\:Corresponding author}$^{\:\:\:1,2,3,4,5}$\\
 $^1$Department of Computer Science and Technology, Tsinghua University, Beijing, China\\
 $^2$Institute for AI Industry Research, Tsinghua University, Beijing, China \\
 $^3$Institute for Artificial Intelligence, Tsinghua University, Beijing, China \\
 $^4$Beijing National Research Center for Information Science and Technology \\
 $^5$Beijing Academy of Artificial Intelligence
}

\date{}

\begin{document}
\maketitle
\begin{abstract}
Word alignment, which aims to align translationally equivalent words between source and target sentences, plays an important role in many natural language processing tasks. Current unsupervised neural alignment methods focus on inducing alignments from neural machine translation models, which does not leverage the full context in the target sequence. In this paper, we propose \textsc{Mask-Align}, a self-supervised word alignment model that takes advantage of the full context on the target side. Our model parallelly masks out each target token and predicts it conditioned on both source and the remaining target tokens. This two-step process is based on the assumption that the source token contributing most to recovering the masked target token should be aligned. We also introduce an attention variant called \textit{leaky attention}, which alleviates the problem of high cross-attention weights on specific tokens such as periods. Experiments on four language pairs show that our model outperforms previous unsupervised neural aligners and obtains new state-of-the-art results.


\end{abstract}

\section{Introduction}
\label{sec:intro}

Word alignment is an important task of finding the correspondence between words in a sentence pair \cite{brown1993mathematics} and used to be a key component of statistical machine translation (SMT) \cite{koehn2003statistical, dyer2013fastalign}. Although word alignment is no longer explicitly modeled in neural machine translation (NMT) \cite{bahdanau14neural, vaswani2017attention}, it is often leveraged to analyze NMT models \cite{tu2016modeling, ding2017visualizing}. Word alignment is also used in many other scenarios such as imposing lexical constraints on the decoding process \cite{arthur2016incorporating, hasler2018neural}, improving automatic post-editing \cite{pal2017neural} , and providing guidance for translators in computer-aided translation \cite{dagan1993robust}.


Compared with statistical methods, neural methods can learn representations end-to-end from raw data and have been successfully applied to supervised word alignment \cite{yang2013word, tamura2014recurrent}. For unsupervised word alignment, however, previous neural methods fail to significantly exceed their statistical counterparts such as \textsc{Fast-Align} \cite{dyer2013fastalign} and \textsc{GIZA++} \cite{och03giza}. 
Recently, there is a surge of interest in NMT-based alignment methods which take alignments as a by-product of NMT systems \cite{li2019word, garg2019jointly, zenkel2019adding, zenkel2020end, chen2020accurate}. Using attention weights or feature importance measures to induce alignments for to-be-predicted target tokens, these methods outperform unsupervised statistical aligners like \textsc{GIZA++} on a variety of language pairs. 


\begin{figure}[t]
\centering
\includegraphics[width=\linewidth]{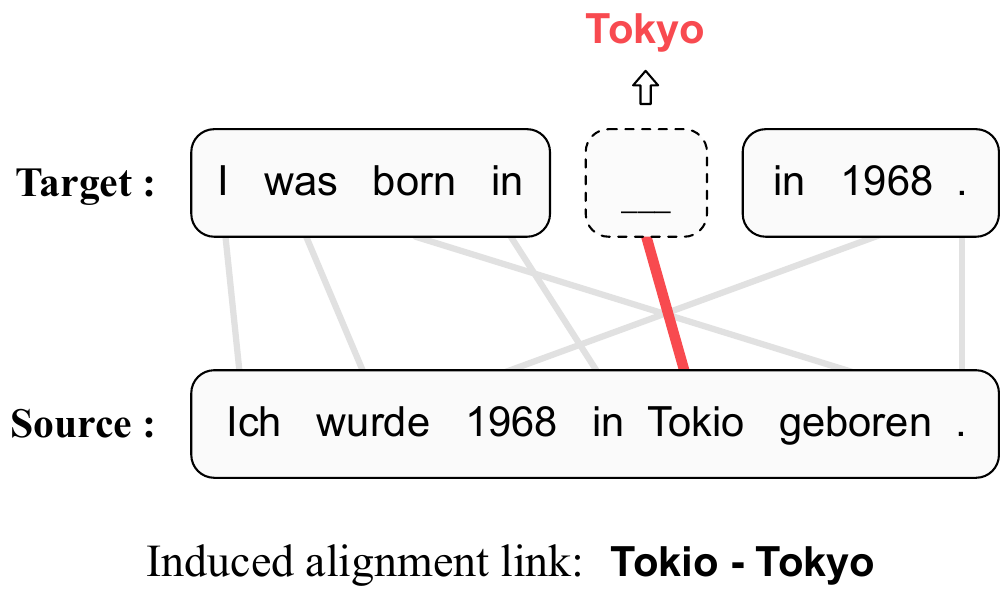}
\caption{An example of inducing an alignment link for target token ``Tokyo'' in \textsc{Mask-Align}. First, we mask out ``Tokyo'' and predict it with source and other target tokens. Then, the source token ``Tokio'' that contributes most to recovering the masked word (highlighted in red) is chosen to be aligned to ``Tokyo''.}
\vspace{-0.3cm}
\label{fig:compare}
\end{figure}

\begin{figure*}[t]
\centering  
\includegraphics[width=0.9\textwidth]{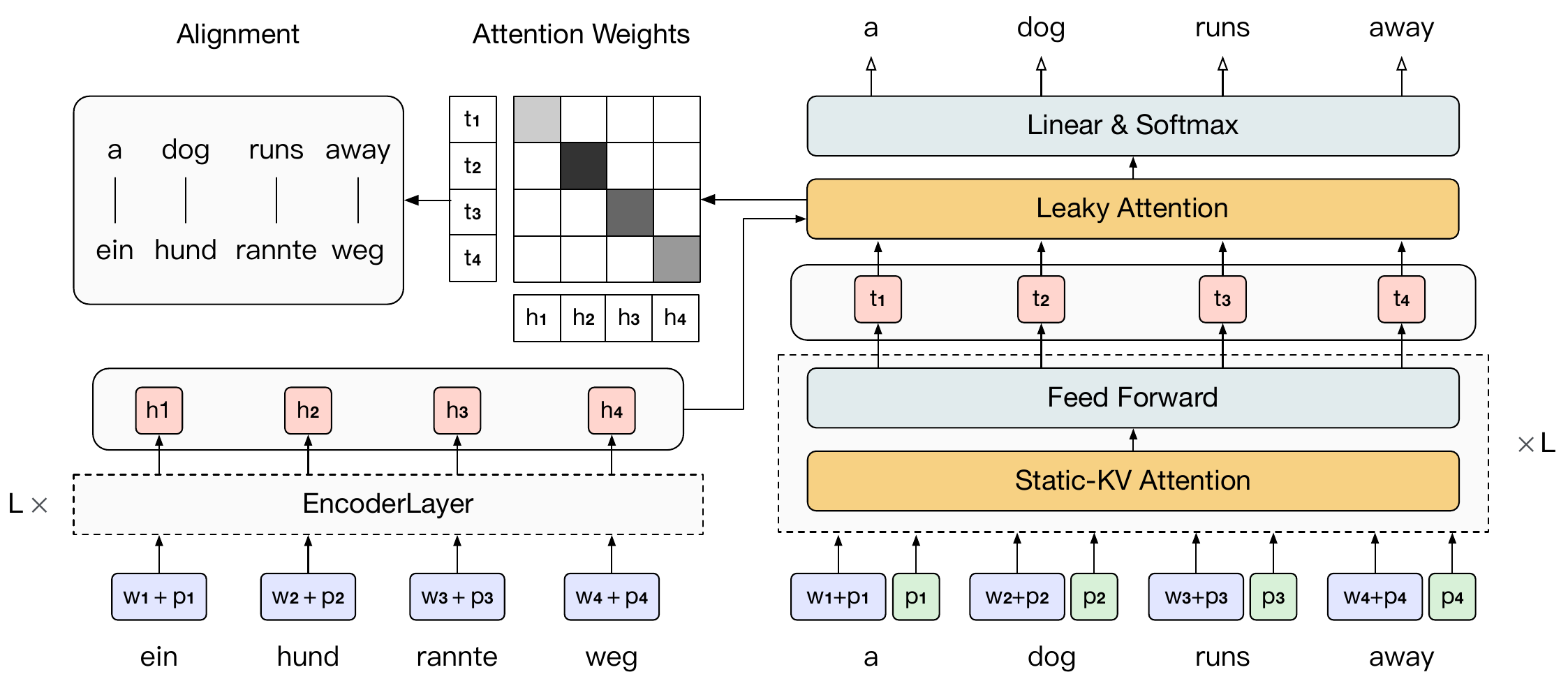}
\caption{The architecture of \textsc{Mask-Align}.}
\label{fig:model}
\end{figure*}

Although NMT-based unsupervised aligners have proven to be effective, they suffer from two major limitations. 
First, due to the autoregressive property of NMT systems \cite{sutskever2014sequence}, they only leverage part of the target context. This inevitably brings noisy alignments when the prediction is ambiguous. Consider the target sentence in Figure \ref{fig:compare}. When predicting ``Tokyo'', an NMT system may generate ``1968'' because future context is not observed, leading to a wrong alignment link (``1968'', ``Tokyo'').
Second, they have to incorporate an additional guided alignment loss \cite{chen2016guided} to outperform \textsc{GIZA++}. This loss requires pseudo alignments of the full training data to guide the training of the model. Although these pseudo alignments can be utilized to partially alleviate the problem of ignoring future context, they are computationally expensive to obtain.


In this paper, we propose a self-supervised model specifically designed for the word alignment task, namely \textsc{Mask-Align}. Our model parallelly masks out each target token and recovers it conditioned on the source and other target tokens. Figure \ref{fig:compare} shows an example where the target token ``Tokyo'' is masked out and re-predicted. Intuitively, as all source tokens except ``Tokio'' can find their counterparts on the target side, ``Tokio'' should be aligned to the masked token. Based on this intuition, we assume that the source token contributing most to recovering a masked target token should be aligned to that 
target token. Compared with NMT-based methods, \textsc{Mask-Align} is able to take full advantage of bidirectional context on the target side and hopefully achieves higher alignment quality. We also introduce an attention variant called \textit{leaky attention} to reduce the high attention weights on specific tokens such as periods. By encouraging agreement between two directional models both for training and inference, our method consistently outperforms the state-of-the-art on four language pairs without using guided alignment loss.

\section{Approach}

Figure \ref{fig:model} shows the architecture of our model. The model predicts each target token conditioned on the source and other target tokens and generates alignments from the attention weights between source and target (Section \ref{sec:modeling}). Specifically, our approach introduces two attention variants, \textit{static-KV attention} and \textit{leaky attention}, to efficiently obtain attention weights for word alignment. To better utilize attention weights from two directions, we encourage agreement between two unidirectional models during both training (Section \ref{sec:training}) and inference (Section \ref{sec:inference}).
    

\subsection{Modeling}
\label{sec:modeling}

Conventional unsupervised neural aligners are based on NMT models \cite{peter2017generating, garg2019jointly}.
Given a source sentence $\mathbf{x} = x_1, \dots, x_J$ and a target sentence $\mathbf{y} = y_1, \dots, y_I$, NMT models the probability of the target
sentence conditioned on the source sentence:
\begin{equation}
    P(\mathbf{y}|\mathbf{x};\theta) = \prod\limits_{i=1}^{I}P(y_{i}|\mathbf{y}_{<i}, \mathbf{x};\theta)
\end{equation}
where $\mathbf{y}_{<i}$ is a partial translation. One problem of this type of approaches is that they fail to exploit the future context on the target side, which is probably helpful for word alignment.


To address this problem, we model the same conditional probability but predict each target token $y_{i}$ conditioned on the source sentence $\mathbf{x}$ and the remaining target tokens $\mathbf{y}\backslash{y_{i}}$:
\begin{equation}
\label{eq:modeling}
    P(\mathbf{y}|\mathbf{x};\theta) = \prod\limits_{i=1}^{I}P(y_{i}|\mathbf{y}\backslash{y_{i}}, \mathbf{x};\theta)
\end{equation}
This equals to masking out each $y_{i}$ and then recovering it. We build our model on top of Transformer \cite{vaswani2017attention} which is the state-of-the-art sequence-to-sequence architecture. Next, we will discuss in detail the implementation of our model.

\subsubsection*{Static-KV Attention}
\label{sec:static-kv attention}

As self-attention is fully-connected, directly computing $\prod_{i=1}^{I}P(y_{i}|\mathbf{y}\backslash{y_{i}}, \mathbf{x};\theta)$ with a vanilla Transformer requires $I$ separate forward passes, in each of which only one target token is masked out and predicted. This is costly and time-consuming. Therefore, how to parallelly mask out and predict all target tokens in a single pass is important.

To do so, a major challenge is to avoid  the representation of a masked token getting involved in the prediction process of itself. Inspired by \citet{kasai2020disco}, we modify the self-attention in the Transformer decoder to perform the forward passes concurrently. Given the word embedding $\mathbf{w}_{i}$ and position embedding $\mathbf{p}_{i}$ for target token $y_{i}$, we first separate the query inputs $\mathbf{q}_{i}$ from key $\mathbf{k}_{i}$ and value inputs $\mathbf{v}_{i}$ to prevent the to-be-predicted token itself from participating in the prediction:
\begin{align}
    \mathbf{q}_{i} &= \mathbf{p}_{i}\mathbf{W}^{Q} \\
    \mathbf{k}_{i} &= (\mathbf{w}_i + \mathbf{p}_{i})\mathbf{W}^{K} \\
    \mathbf{v}_{i} &= (\mathbf{w}_i + \mathbf{p}_{i})\mathbf{W}^{V}
\end{align}
where $\mathbf{W}^{Q}$, $\mathbf{W}^{K}$ and $\mathbf{W}^{V}$ are parameter matrices.
The hidden representation $\mathbf{h}_{i}$ for $y_{i}$ is computed by attending to keys and values, $\mathbf{K}_{\neq{i}}$ and $\mathbf{V}_{\neq{i}}$, that correspond to the remaining tokens $\mathbf{y}\backslash{y_{i}}$:
\begin{align}
    \mathbf{h}_{i} &= \mathrm{Attention}(\mathbf{q}_{i}, \mathbf{K}_{\neq{i}}, \mathbf{V}_{\neq{i}}) \\
    \mathbf{K}_{\neq{i}} &= \mathrm{Concat}(\{\mathbf{k}_{m}|m\neq{i}\}) \\
	\mathbf{V}_{\neq{i}} &= \mathrm{Concat}(\{\mathbf{v}_{m}|m\neq{i}\})
\end{align}
In this way, we ensure that $\mathbf{h}_{i}$ is isolated from the word embedding $\mathbf{w}_{i}$ in a single decoder layer. However, there exists a problem of information leakage if we update the key and value inputs for each position across decoder layers since they will contain the representation of each position from previous layers. Therefore, we keep the key and value inputs unchanged and only update the query inputs to avoid information leakage:
\begin{align}
	\mathbf{h}_{i}^{l} &= \mathrm{Attention}(\mathbf{q}_{i}^{l}, \mathbf{K}_{\neq{i}}, \mathbf{V}_{\neq{i}}) \\
	\mathbf{q}_{i}^{l} &= \mathbf{h}_{i}^{l-1}\mathbf{W}^{Q}
\end{align}
where $\mathbf{q}_{i}^{l}$ and $\mathbf{h}_{i}^{l}$ denote the query inputs and hidden states for $y_{i}$ in the $l$-th layer, respectively. $\mathbf{h}_{i}^{0}$ is initialized with $\mathbf{p}_{i}$. We name this variant of attention the \textbf{static-KV attention}. By static-KV, we mean the keys and values are unchanged across different layers in our approach.
Our model replaces all self-attention in the decoder with static-KV attention.

\subsubsection*{Leaky Attention}
\label{sec:leaky_attention}


\begin{figure}[t]
\centering
\includegraphics[width=0.9\linewidth]{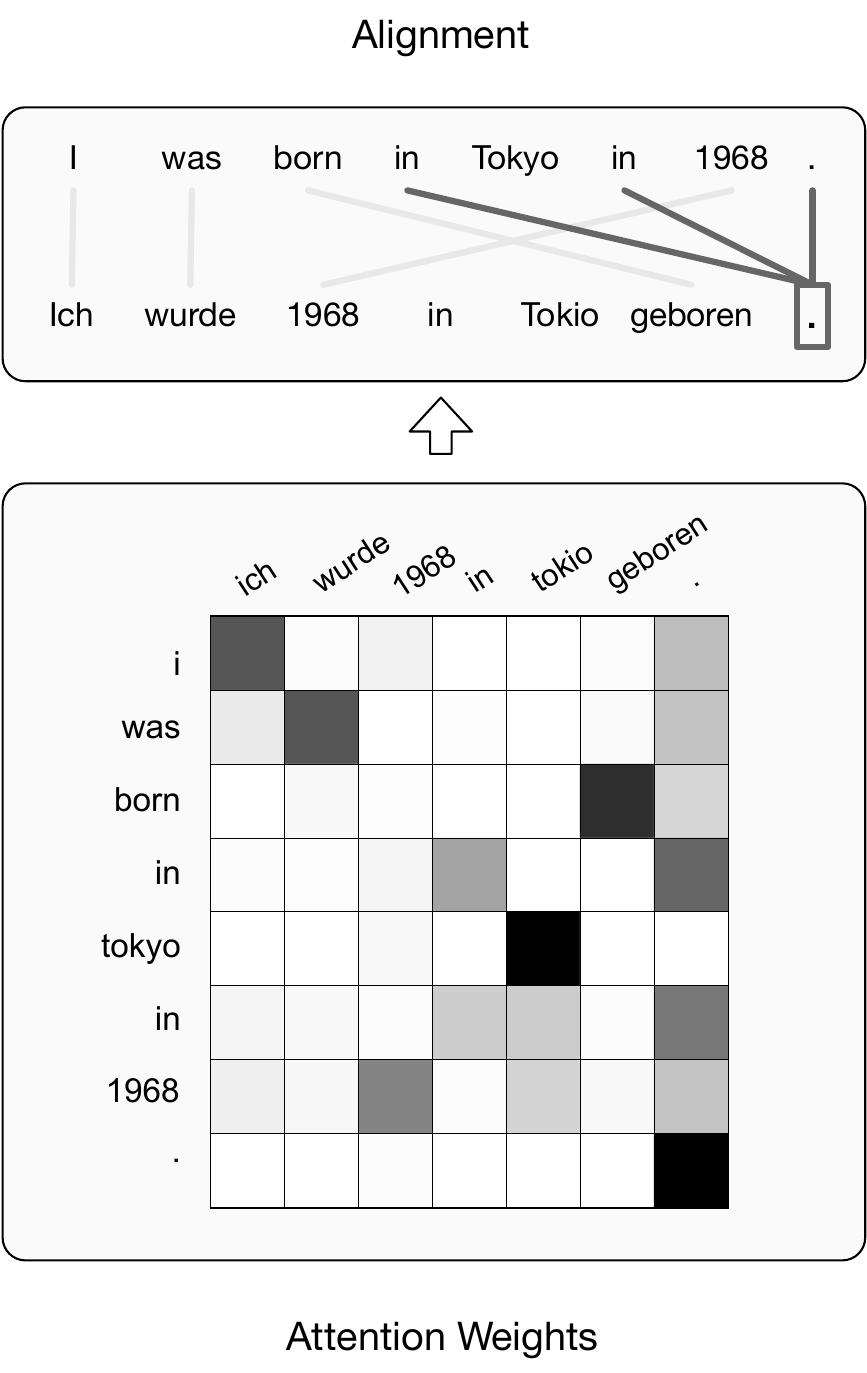}
\caption{An example of inducing alignments from attention weights where the source token ``.'' has high attention weights. The two ``in''s in the target sentence are wrongly aligned to ``.'' because of the high attention weights on it.}
\vspace{-0.5cm}
\label{fig:garbage}
\end{figure}

Extracting alignments from vanilla cross-attention often suffers from the high attention weights on some specific source tokens such as periods, [EOS], or other high frequency tokens (see Figure \ref{fig:garbage}). This is similar to the ``garbage collectors'' effect \cite{moore2004improving} in statistical aligners, where a source token is aligned to too many target tokens.
Hereinafter, we will refer to these tokens as \textit{collectors}. As a result of such effect, many target tokens (e.g., the two ``in''s in Figure \ref{fig:garbage}) will be incorrectly aligned to the collectors according to the attention weights.

\begin{figure}[t]
\centering
\includegraphics[width=1.0\linewidth]{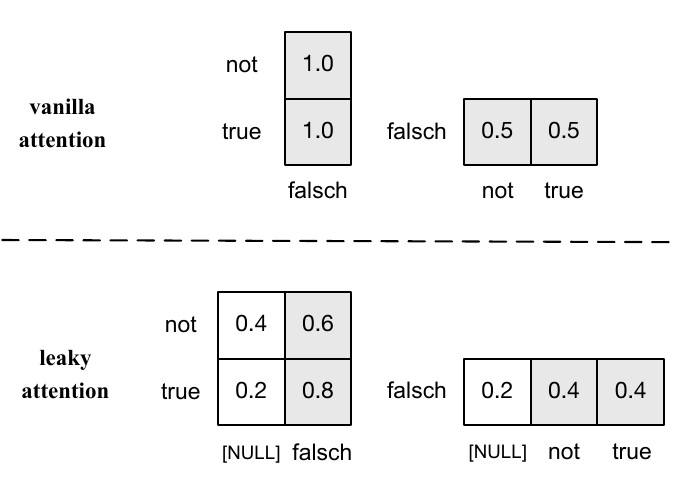}
\caption{An illustrative example of the attention weights from two directional models using vanilla and leaky attention. Leaky attention provides a leak position ``[NULL]'' to collect extra attention weights.}
\label{fig:agreement}
\end{figure}

This phenomenon has been studied in previous works \cite{clark2019does, kobayashi2020attention}. \citet{kobayashi2020attention} show that the norms of the value vectors for the collectors are usually small, making their influence on attention outputs actually limited. 
We conjecture that this phenomenon is due to the incapability of NMT-based aligners to deal with tokens that have no counterparts on the other side because there is no empty (NULL) token that is widely used in statistical aligners \cite{brown1993mathematics, och03giza}.

We propose to explicitly model the NULL token with an attention variant, namely \textbf{leaky attention}. As shown in Figure \ref{fig:agreement}, when calculating cross-attention weights, leaky attention provides an extra ``leak'' position in addition to the encoder outputs. Acting as the NULL token, this leak position is expected to address the biased attention weight problem.
To be specific, we parameterize the key and value vectors as $\mathbf{k}_{\text{NULL}}$ and $\mathbf{v}_{\text{NULL}}$ for the leak position in the cross-attention, and concatenate them with the transformed vectors of the encoder outputs. The attention output $\mathbf{z}_{i}$ is computed as follows: 
\begin{align}
	\mathbf{z}_{i} &= \mathrm{Attention}(\mathbf{h}_{i}^{L}\mathbf{W}^{Q}, \mathbf{K}, \mathbf{V}) \\
    \mathbf{K} &= \mathrm{Concat}(\mathbf{k}_{\text{NULL}}, \mathbf{H}_{\text{enc}}\mathbf{W}^{K}) \\
	\mathbf{V} &= \mathrm{Concat}(\mathbf{v}_{\text{NULL}}, \mathbf{H}_{\text{enc}}\mathbf{W}^{V})
\end{align}
where $\mathbf{H}_{\text{enc}}$ denotes encoder outputs. \footnote{A similar attention implementation can be found in \url{https://github.com/pytorch/fairseq/blob/master/fairseq/modules/multihead_attention.py}.} We use a normal distribution with a mean of 0 and a small deviation to initialize $\mathbf{k}_{\text{NULL}}$ and $\mathbf{v}_{\text{NULL}}$ to ensure that their initial norms are rather small. When extracting alignments, we only consider the attention matrix without the leak position. 

Note that leaky attention is different from adding a special token in the source sequence, which will share the same high attention weights with the existing collector instead of calibrating it \cite{vig2019analyzing}. Our parameterized method is more flexible than Leaky-Softmax \cite{sabour2017dynamic} which adds an extra dimension with the value of zero to the routing logits. In Section \ref{sec:training}, we will show that leaky attention is also helpful for applying agreement-based training on two directional models.

We remove the cross-attention in all but the last decoder layer. This makes the interaction between the source and target restricted in the last layer. Our experiments demonstrate that this modification improves alignment results with fewer model parameters.

\subsection{Training}
\label{sec:training}

To better utilize the attention weights from two directions, we apply an agreement loss in the training process to improve the symmetry of our model, which has proven effective in statistical alignment models \cite{liang2006alignment, liu2015generalized}. 
Given a parallel sentence pair $\left \langle \mathbf{x}, \mathbf{y} \right \rangle$, we can obtain the attention weights from two different directions, denoted as $\boldsymbol{W}_{\mathbf{x\rightarrow{y}}}$ and $\boldsymbol{W}_{\mathbf{y\rightarrow{x}}}$. As alignment is bijective, $\boldsymbol{W}_{\mathbf{x\rightarrow{y}}}$ is supposed to be equal to the transpose of $\boldsymbol{W}_{\mathbf{y\rightarrow{x}}}$. We encourage this kind of symmetry through an agreement loss: \begin{equation}
    \mathcal{L}_{a} = \mathrm{MSE}\left(\boldsymbol{W}_{\mathbf{x\rightarrow{y}}}, \boldsymbol{W}_{\mathbf{y\rightarrow{x}}}^{\top}\right)
\end{equation} 
where $\mathrm{MSE}$ represents the mean squared error.


For vanilla attention, $\mathcal{L}_{a}$ is hardly small because of the normalization constraint. As shown in Figure \ref{fig:agreement}, due to the use of softmax activation, the minimal value of $\mathcal{L}_{a}$ is $0.25$ for vanilla attention. Using leaky attention, our approach can achieve a lower agreement loss ($\mathcal{L}_{a}$ = 0.1) by adjusting the weights on the leak position.

However, our model may converge to a degenerate case of zero agreement loss where attention weights are all zero except for the leak position. We circumvent this case by introducing an entropy loss on the attention weights:
\begin{align}
    \mathcal{L}_{e, \mathbf{x\rightarrow{y}}} &= -\frac{1}{I}\sum\limits_{i=1}^{I}\sum\limits_{j=1}^{J}\tilde{W}^{ij}_{\mathbf{x\rightarrow{y}}}\log{\tilde{W}_{ij}} \\
    \tilde{W}^{ij}_{\mathbf{x\rightarrow{y}}} &= \frac{W^{ij}_{\mathbf{x\rightarrow{y}}} + \lambda}{\sum_{j}{(W^{ij}_{\mathbf{x\rightarrow{y}}}+\lambda)}} \label{eq:renorm}
\end{align}
where $\tilde{W}^{ij}_{\mathbf{x\rightarrow{y}}}$ is the renormalized attention weights and $\lambda$ is a smoothing hyperparamter. Similarly, we have $ \mathcal{L}_{e, \mathbf{y\rightarrow{x}}}$ for the inverse direction.

We jointly train two directional models using the following loss:
\begin{equation}
\begin{aligned}
    \mathcal{L} = \,\,&\mathcal{L}_{\mathbf{x\rightarrow{y}}} + \mathcal{L}_{\mathbf{y\rightarrow{x}}} + \alpha\mathcal{L}_{a} + \\ &\beta(\mathcal{L}_{e, \mathbf{x\rightarrow{y}}} + \mathcal{L}_{e, \mathbf{y\rightarrow{x}}})
\label{eq:loss}
\end{aligned}
\end{equation}
where $\mathcal{L}_{\mathbf{x\rightarrow{y}}}$ and $\mathcal{L}_{\mathbf{y\rightarrow{x}}}$ are NLL losses, $\alpha$ and $\beta$ are hyperparameters. 

\subsection{Inference}
\label{sec:inference}

When extracting alignments, we compute an alignment score $S_{ij}$ for $y_{i}$ and $x_{j}$ as the harmonic mean of attention weights $W_{\mathbf{x\rightarrow{y}}}^{ij}$ and $W_{\mathbf{y\rightarrow{x}}}^{ji}$ from two directional models:
\begin{equation}
    S_{ij} = \frac{2\,W_{\mathbf{x\rightarrow{y}}}^{ij}\,W^{ji}_{\mathbf{y\rightarrow{x}}}}{W_{\mathbf{x\rightarrow{y}}}^{ij}+W^{ji}_{\mathbf{y\rightarrow{x}}}}
\end{equation}
We use the harmonic mean because we assume a large $S_{ij}$ requires both $W_{\mathbf{x\rightarrow{y}}}^{ij}$ and $W_{\mathbf{y\rightarrow{x}}}^{ji}$ to be large. Word alignments can be induced from the alignment score matrix as follows:
\begin{equation}
\label{eq:inference}
A_{i j}=\left\{\begin{array}{ll}1 & \text { if } S_{ij} \geq{\tau} \\ 0 & \text { otherwise }\end{array}\right.
\end{equation}
where $\tau$ is a threshold.

\begin{table*}[t]
    \centering
    \begin{tabular}{lccccc}
        \toprule
        \textbf{Method} & \textbf{Guided} & \textbf{De-En} & \textbf{En-Fr} & \textbf{Ro-En} & \textbf{Zh-En} \\
        \midrule
        \textsc{Fast-Align} \cite{dyer2013fastalign} & N & 25.7 & 12.1  & 31.8 & - \\
        \textsc{GIZA++} \cite{och03giza}  & N & 17.8 & \,\,\,6.1 & 26.0 & 18.5 \\
        \midrule
        \textsc{Naive-Att} \cite{garg2019jointly}  & N & 31.9 & 18.5 & 32.9 & 28.9 \\
        \textsc{Naive-Att-Last} & N & 28.4 & 17.7 & 32.4 & 26.4 \\
        \textsc{AddSGD} \cite{zenkel2019adding}  & N & 21.2 & 10.0 & 27.6 &   -                \\
        \textsc{Mtl-Fullc} \cite{garg2019jointly}  & N & 20.2 & \,\,\,7.7  & 26.0 &   -                \\
        \textsc{BAO} \cite{zenkel2020end} & N & 17.9 & \,\,\,8.4  & 24.1 &   -                \\
        \textsc{Shift-Att} \cite{chen2020accurate} & N & 17.9 & \,\,\,6.6  & 23.9 & 20.2                  \\ 
        \midrule
        \textsc{Mtl-Fullc-GZ} \cite{garg2019jointly} & Y & 16.0 & \,\,\,4.6 & 23.1 & - \\
        \textsc{BAO-Guided} \cite{zenkel2020end} & Y & 16.3 & \,\,\,5.0 & 23.4 & - \\
        \textsc{Shift-AET} \cite{chen2020accurate} & Y & 15.4 & \,\,\,4.7 & 21.2 & 17.2 \\
        \midrule
        \textsc{Mask-Align} & N          & \textbf{14.4} & \,\,\,\textbf{4.4} & \textbf{19.5} & \textbf{13.8} \\
        \bottomrule
    \end{tabular}
    \caption{Alignment Error Rate (AER) scores on four datasets for different alignment methods. The lower AER, the better. ``Guided'' denotes whether the guided alignment loss is used during training. All results are symmetrized. We highlight the best results for each language pair in bold.}
    \label{tb:main}
\end{table*}

\section{Experiments}

\subsection{Datasets}

We conducted our experiments on four public datasets: German-English (De-En), English-French (En-Fr), Romanian-English (Ro-En) and Chinese-English (Zh-En). The Chinese-English training set is from the LDC corpus that consists of 1.2M sentence pairs. For validation and testing, we used the Chinese-English alignment dataset from \citet{liu2005log}\footnote{\url{http://nlp.csai.tsinghua.edu.cn/~ly/systems/TsinghuaAligner/TsinghuaAligner.html}}, which contains 450 sentence pairs for validation and 450 for testing. For other three language pairs, we followed the experimental setup in \cite{zenkel2019adding,zenkel2020end} and used the preprocessing scripts from \citet{zenkel2019adding}\footnote{\url{https://github.com/lilt/alignment-scripts}}. Following \citet{ding2019saliency}, we take the last 1000 sentences of the training data for these three datasets as validation sets. We used a joint source and target Byte Pair Encoding (BPE) \cite{sennrich2016bpe} with 40k merge operations. During training, we filtered out sentences with the length of 1 to ensure the validity of the masking process.

\subsection{Settings}

We implemented our model based on the Transforemr architecture \cite{vaswani2017attention}. The encoder consists of 6 standard Transformer encoder layers. The decoder is composed of 6 layers, each of which contains static-KV attention while only the last layer is equipped with leaky attention. We set the embedding size to 512, the hidden size to 1024, and attention heads to 4. The input and output embeddings are shared for the decoder.

We trained the models with a batch size of 36K tokens. We used early stopping based on the pre-diction accuracy on the validation sets. We tuned the hyperparameters via grid search on the Chinese-English validation set as it contains gold word alignments. In all of our experiments, we set $\lambda=0.05$ (Eq. (\ref{eq:renorm})), $\alpha=5$, $\beta=1$ (Eq. (\ref{eq:loss})) and $\tau = 0.2$ (Eq. (\ref{eq:inference})). 
The evaluation metric is Alignment Error Rate (AER) \cite{och2000improved}.

\subsection{Baselines}

We introduce the following unsupervised neural baselines besides two statistical baselines \textsc{Fast-Align} and \textsc{GIZA++}:
\begin{itemize}[topsep=5pt]
\setlength{\itemsep}{2pt}
\setlength{\parsep}{0pt}
\setlength{\parskip}{0pt}
    \item \textsc{Naive-Att} \cite{garg2019jointly}: a method that induces alignments from cross-attention weights of the best (usually penultimate) decoder layer in a vanilla Tranformer.
    \item \textsc{Naive-Att-Last}: same as \textsc{Naive-Att} except that only the last decoder layer performs cross-attention.
    \item \textsc{AddSGD} \cite{zenkel2019adding}: a method that adds an extra alignment layer to repredict the to-be-aligned target token.
    \item \textsc{Mtl-Fullc} \cite{garg2019jointly}: a method that supervises an attention head with symmetrized \textsc{Naive-Att} alignments in a multi-task learning framework.
    \item \textsc{BAO} \cite{zenkel2020end}: an improved version of \textsc{AddSGD} that extracts alignments with Bidirectional Attention Optimization.
    \item \textsc{Shift-Att} \cite{chen2020accurate}: a method that induces alignments when the to-be-aligned tatget token is the decoder input instead of the output.
\end{itemize}

We also included three additional baselines with guided training: (1) \textsc{Mtl-Fullc-GZ} \cite{garg2019jointly} which replaces the alignment labels in \textsc{Mtl-Fullc} with GIZA++ results, (2) \textsc{BAO-Guided} \cite{zenkel2020end} which uses alignments from \textsc{BAO} for guided alignment training, (3) \textsc{Shift-AET} \cite{chen2020accurate} which trains an additional alignment module with supervision from symmetrized \textsc{Shift-Att} alignments.

\subsection{Main Results}

Table \ref{tb:main} shows the results on four datasets. Our approach significantly outperforms all statistical and neural baselines. Specifically, it improves over GIZA++ by 1.7-6.5 AER points across different language pairs without using any guided alignment loss, making it a good substitute to this commonly used statistical alignment tool. Compared to \textsc{Shift-Att}, the best neural methods without guided training, our approach achieves a gain of 2.2-6.4 AER points with fewer parameters (as we remove some cross-attention sublayers in the decoder).

When compared with baselines using guided training, we find \textsc{Mask-Align} still achieves substantial improvements over all methods. For example, on the Romanian-English dataset, it improves over \textsc{Shift-AET} by 1.7 AER points. Recall that our method is fully end-to-end, which does not require a time-consuming process of obtaining pseudo alignments for full training data.


\begin{figure*}[t]
\centering  
\subfigcapskip=5pt
\subfigure[Vanilla Attention]{
\begin{minipage}{7cm}
\centering
\includegraphics[width=\textwidth]{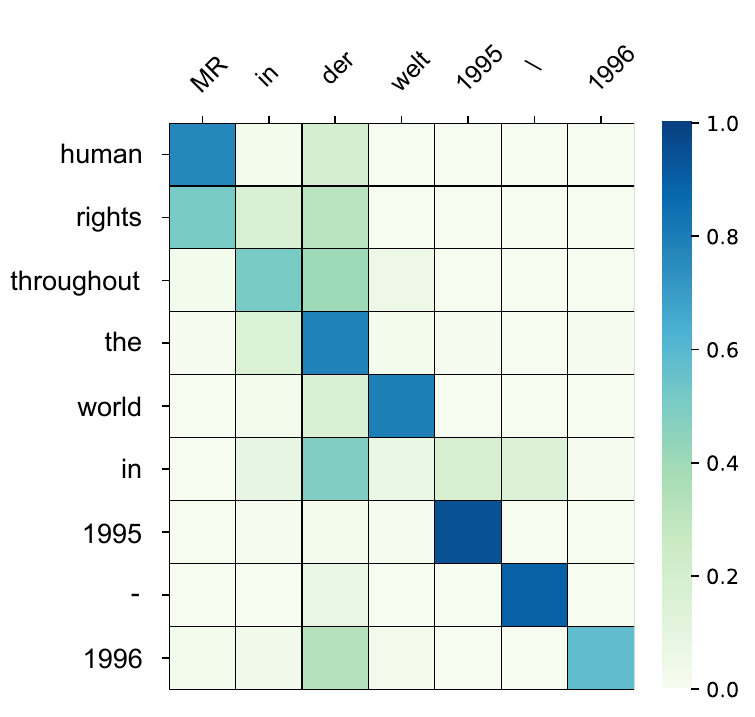}
\end{minipage}
}
\subfigure[Leaky Attention]{
\begin{minipage}{7cm}
\centering
\includegraphics[width=\textwidth]{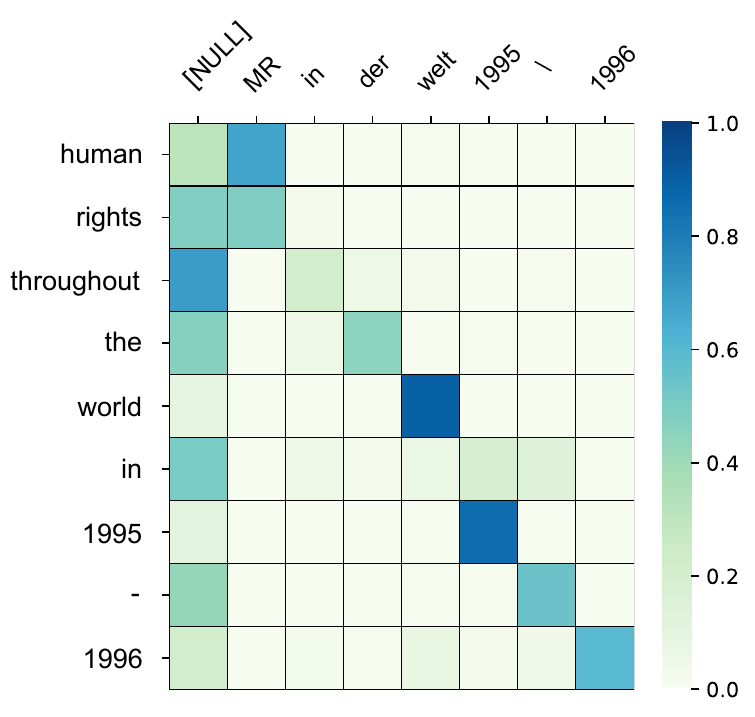}
\end{minipage}
}
\caption{Attention weights from vanilla and leaky attention. ``MR'' is short for ``menschenrechte'', which means ``human rights'' in English. We use ``\texttt{[NULL]}'' to denote the leak position.}
\label{fig:leaky_exp}
\end{figure*}

\begin{table}[t]
\centering
 \begin{tabular}{cccc}
  \toprule
  Masked & Leaky & Agree & AER\\
  \midrule
  \xmark & \xmark & \xmark & 28.4 \\
  \cmark & \xmark & \xmark & 27.2 \\
  \xmark & \cmark & \xmark & 28.3 \\
  \xmark & \xmark & \cmark & 26.6 \\
  \xmark & \cmark & \cmark & 23.4 \\
  \cmark & \xmark & \cmark & 17.6 \\
  \cmark & \cmark & \xmark & 17.2 \\
  \midrule
  \cmark & \cmark & \cmark & \textbf{14.4} \\
  \bottomrule
 \end{tabular}
 \caption{Ablation study on the German-English dataset. We use ``Masked'' to denote the masked modeling with static-KV attention in Section \ref{sec:static-kv attention}, ``Leaky'' to denote the leaky attention in Section \ref{sec:leaky_attention} and ``Agree'' to denote the agreement-based training and inference in Sections \ref{sec:training} and \ref{sec:inference}.} 
 \label{tb:ablation}
 \vspace{-1em}
\end{table}

\begin{table*}[t]
    \centering
     \begin{tabular}{rp{1.4cm}<{\centering}p{1.1cm}<{\centering}p{1.1cm}<{\centering}p{1cm}<{\centering}p{1cm}<{\centering}p{1cm}<{\centering}p{1cm}<{\centering}p{1cm}<{\centering}}
      \toprule
      source sentence & \texttt{[NULL]} & \texttt{MR} & \texttt{in} & \texttt{der} & \texttt{welt} & \texttt{1995} & \texttt{$\backslash$} & \texttt{1996} \\
      \midrule
     vanilla attention & - & 21.1 & 11.7 & \textbf{5.2} & 15.0 & 21.2 & 17.7 & 21.8 \\
     leaky attention & \textbf{1.9} & 28.5 & 17.2 & 18.1 & 20.2 & 24.2 & 21.4 & 23.8 \\
      \bottomrule
     \end{tabular}
     \caption{Norms of the transformed value vectors of different source tokens in Figure \ref{fig:leaky_exp}. We mark the minimum norm for each variant of attention with boldface.}
     \label{tb:leakyattn}
\end{table*}

\subsection{Ablation Study}


Table \ref{tb:ablation} shows the ablation results on the German-English dataset. As we can see, masked modeling seems to play a critical role since removing it will deteriorate the performance by at least 9.0 AER. We also find that leaky attention and agreement-based training and inference are both important. Removing any of them will significantly diminish the performance.




\subsection{Effect of Leaky Attention}

Figure \ref{fig:leaky_exp} shows the attention weights from vanilla and leaky attention and Table \ref{tb:leakyattn} presents the norms of the transformed value vectors of each source token for two types of attention. For vanilla attention, we can see large weights on the high frequency token ``der'' and the small norm of its transformed value vector. As a result, the target token ``in'' will be wrongly aligned to ``der''. While for leaky attention, we observe a similar phenomenon on the leak position ``\texttt{[NULL]}'', and ``in'' will not be aligned to any source tokens since the weights on all source tokens are small. This example shows leaky attention can effectively prevent the collector phenomenon.

\subsection{Analysis}

\noindent \textbf{Removing End Punctuation} To further investigate the performance of leaky attention, we tested an extraction method that excludes the attention weights on the end punctuation of a source sentence. The reason behind this is that when the source sentence contains the end punctuation, it will act as the collector in most cases. Therefore removing it will alleviate the effect of collectors to a certain extent. Table \ref{tb:punc} shows the comparison results. For vanilla attention, removing end punctuation obtains a gain of 7.7 AER points.
For leaky attention, however, such extraction method brings no improvement on alignment quality. This suggests that leaky attention can effectively alleviate the problem of collectors.

\begin{table}
\centering
 \begin{tabular}{rcc}
  \toprule
  Method & w/ punc. & w/o punc.\\
  \midrule
  vanilla attention & 27.2 & 17.7 \\
  leaky attention & 17.2 & 17.4 \\
  \bottomrule
 \end{tabular}
 \caption{Comparison of AER with and without considering the attention weights on end punctuation.}
 \label{tb:punc}
\end{table}

\begin{figure*}[t]
\centering  
\subfigure[Reference]{
\begin{minipage}{3.8040cm}
\centering
\includegraphics[width=\textwidth]{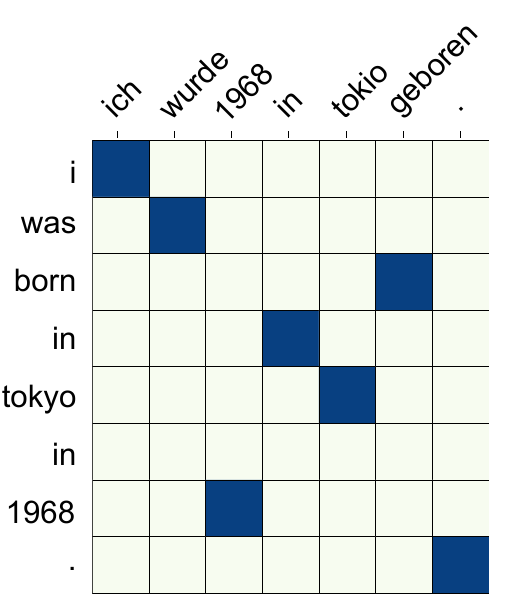}
\end{minipage}
\label{fig:case1}
}
\subfigure[\textsc{Naive-Att-Last}]{
\begin{minipage}{3.8880cm}
\centering
\includegraphics[width=\textwidth]{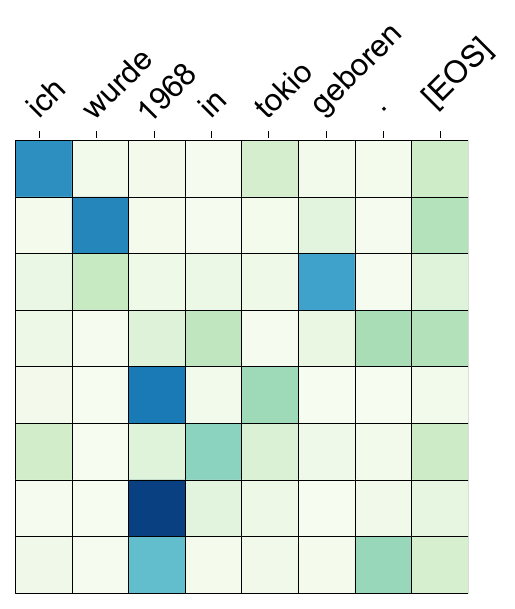}
\end{minipage}
\label{fig:case2}
}
\subfigure[\textsc{Shift-Att}]{
\begin{minipage}{3.8640cm}
\centering
\includegraphics[width=\textwidth]{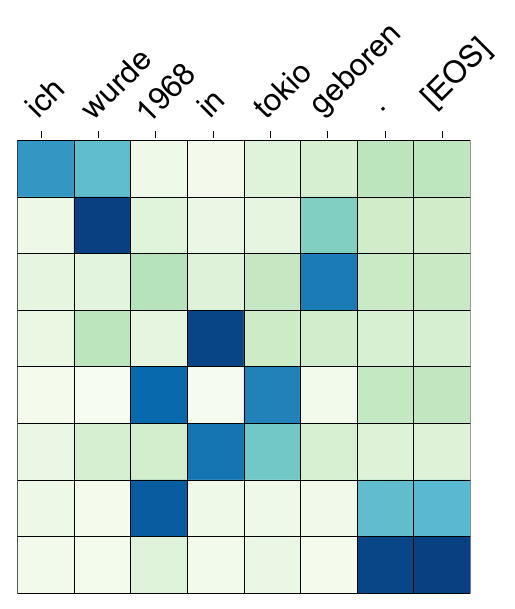}
\end{minipage}
\label{fig:case3}
}
\subfigure[\textsc{Mask-Align}]{
\begin{minipage}{3.2280cm}
\centering
\includegraphics[width=\textwidth]{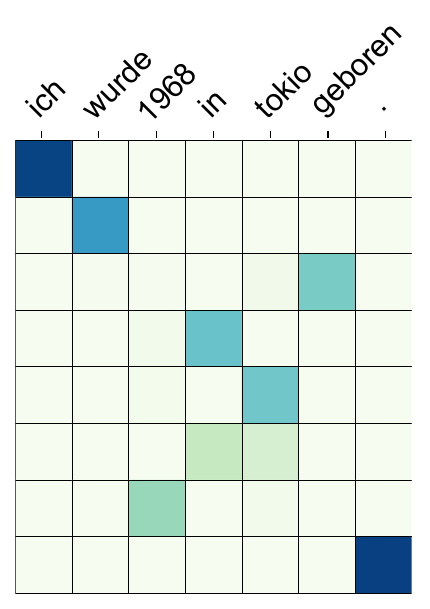}
\end{minipage}
\label{fig:case4}
}
\caption{Attention weights from different models for the example in Figure \ref{fig:compare}. Gold alignment is shown in (a). For target token ``tokyo'', NMT-based methods \textsc{Naive-Att-Last} (b) and \textsc{Shift-Att} (c) assign high weights to the wrongly aligned source token ``1968'', while \textsc{Mask-Align} (d) focuses on the correct source token ``tokio''.}
\label{fig:case}
\end{figure*}

\begin{figure}[t]
\centering
\includegraphics[width=0.485\textwidth]{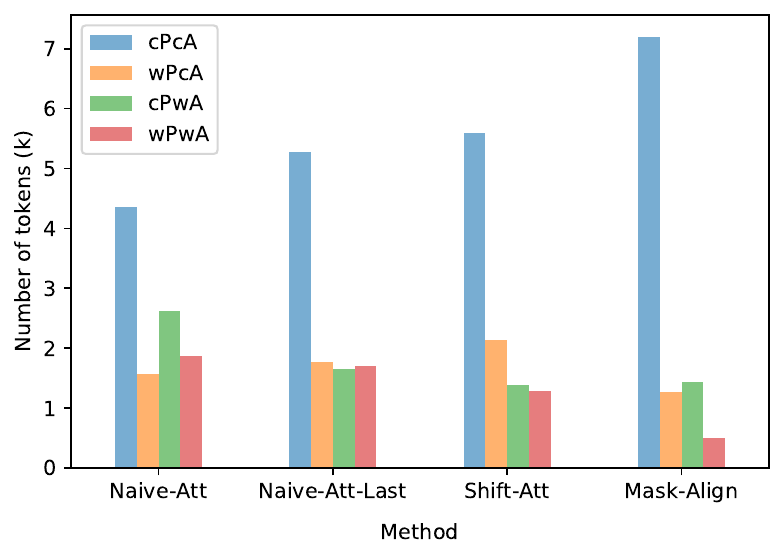}
\caption{Relations between prediction and alignment for different methods.}
\vspace{-0.5cm}
\label{fig:pred_align}
\end{figure}

\vspace{0.5em}

\noindent \textbf{Case Study} Figure \ref{fig:case} shows the attention weights from four different models for the example in Figure \ref{fig:compare}. As we have discussed in Section \ref{sec:intro}, in this example, NMT-based methods might fail to resolve ambiguity when predicting the target token ``tokyo''. From the attention weight matrices, we can see that NMT-based methods (Figures \ref{fig:case2} and \ref{fig:case3}) indeed put high weights wrongly on ``1968'' in the source sentence.
As for \textsc{Mask-Align}, we can see that the attention weights are highly consistent with the gold alignment, showing that our method can generate sparse and accurate attention weights.
\vspace{0.5em}

\noindent \textbf{Prediction and Alignment} We analyzed the relevance between the correctness of word-level prediction and alignment. We regard a word as correctly predicted if any of its subwords are correct and as correctly aligned if one of its possible alignment is matched. Figure \ref{fig:pred_align} shows the results. We divide target tokens into four categories:
\begin{enumerate}[parsep=0em]
    \item cPcA: correct prediction \& correct alignment;
    \item wPcA: wrong prediction \& correct alignment;
    \item cPwA: correct prediction \& wrong alignment;
    \item wPwA: wrong prediction \& wrong alignment.
\end{enumerate}

Compared with other methods, \textsc{Mask-Align} significantly reduces the alignment errors caused by wrong predictions (wPwA). In addition, the number of the tokens with correct prediction but wrong alignment (cPwA) maintains at a low level, indicating that our model does not degenerate into a target masked language model despite the use of bidirectional target context.

\section{Related Work}

Our work is closely related to unsupervised neural word alignment. While early unsupervised neural aligners \cite{tamura2014recurrent, alkhouli2016alignment, peter2017generating} failed to outperform their statistical counterparts such as \textsc{Fast-Align} \cite{dyer2013fastalign} and \textsc{GIZA++} \cite{och03giza}, recent studies have made significant progress by inducing alignments from NMT models  \cite{garg2019jointly, zenkel2019adding, zenkel2020end, chen2020accurate}. 
Our work differs from prior studies in that we design a novel self-supervised model that is capable of utilizing more target context than NMT-based models to generate high quality alignments without using guided training.

Our work is also inspired by the success of conditional masked language models (CMLMs) \cite{ghazvininejad2019mask}, which have been applied to non-autoregressive machine translation. The CMLM can leverage both previous and future context on the target side for sequence-to-sequence tasks with the masking mechanism. \citet{kasai2020disco} extend it with a disentangled context Transformer that predicts every target token conditioned on arbitrary context. By taking the characteristics of word alignment into consideration, we propose to use static-KV attention to achieve masking and aligning in parallel.
To the best of our knowledge, this is the first work that incorporates a CMLM into alignment models.

\section{Conclusion}

We have presented a self-supervised neural alignment model \textsc{Mask-Align}. Our model parallelly masks out and predicts each target token. We propose static-KV attention and leaky attention to achieve parallel computation and address the ``garbage collectors'' problem, respectively. Experiments show that \textsc{Mask-Align} achieves new state-of-the-art results without using the guided alignment loss. In the future, we plan to extend our method to directly generate symmetrized alignments without leveraging the agreement between two unidirectional models.

\section*{Acknowledgments}

This work was supported by the National Key R\&D Program of China (No. 2017YFB0202204), National Natural Science Foundation of China (No.61925601, No. 61772302) and Huawei Noah’s Ark Lab.
We thank all anonymous reviewers for their valuable comments and suggestions on this work.

\bibliography{acl2021}
\bibliographystyle{acl_natbib}
\end{document}